\documentclass[runningheads]{llncs}
\usepackage{graphicx}
\usepackage{amsmath,amssymb} 
\usepackage{color}
\usepackage{subfig}

\usepackage{algorithm}
\usepackage{algorithmic}

\DeclareMathOperator*{\argmax}{arg\,max}

\newcommand{\x}{\mathbf{x}}
\newcommand{\y}{\mathbf{y}}

\graphicspath{{fig/}}

\begin{document}

\newcommand{\point}{
    \raise0.7ex\hbox{.}
    }


\pagestyle{headings}

\def\ACCV14SubNumber{72}  

\mainmatter


\title{A Latent Clothing Attribute Approach for Human Pose Estimation} 

\titlerunning{A Latent Clothing Attribute Approach for Human Pose Estimation} 

\authorrunning{Weipeng Zhang, Jie Shen, Guangcan Liu, Yong Yu} 

\author{Weipeng Zhang$^{\dag}$, Jie Shen$^{\dag}$, Guangcan Liu$^{\ddag}$, Yong Yu$^{\dag}$} 
\institute{$^{\dag}$Shanghai Jiao Tong University, $^{\ddag}$Nanjing University of Information Science and Technology} 

\maketitle

\begin{abstract}
As a fundamental technique that concerns several vision tasks such as image parsing, action recognition and clothing retrieval, human pose estimation (HPE) has been extensively investigated in recent years. To achieve accurate and reliable estimation of the human pose, it is well-recognized that the clothing attributes are useful and should be utilized properly. Most previous approaches, however, require to manually annotate the clothing attributes and are therefore very costly. In this paper, we shall propose and explore a \emph{latent} clothing attribute approach for HPE. Unlike previous approaches, our approach models the clothing attributes as latent variables and thus requires no explicit labeling for the clothing attributes. The inference of the latent variables are accomplished by utilizing the framework of latent structured support vector machines (LSSVM). We employ the strategy of \emph{alternating direction} to train the LSSVM model: In each iteration, one kind of variables (e.g., human pose or clothing attribute) are fixed and the others are optimized. Our extensive experiments on two real-world benchmarks show the state-of-the-art performance of our proposed approach.
\end{abstract}

\section{Introduction}

Human oriented technology has a central role in computer vision and can greatly advance daily-life related applications. For example, face verification for surveillance~\cite{face} and clothing parsing for fashion search~\cite{clothliu}. One of the most fundamental human oriented techniques is the well-known \emph{human pose estimation} (HPE) in 2D images. In general, HPE could facilitate many applications, e.g., action recognition~\cite{action}, image segmentation~\cite{ladicky}, etc. However, it is difficult to accurately estimate the human pose in unconstrained environments, especially in the presence of vision occlusions and background clutters.

\begin{figure}[tbp]
\centering
\subfloat[]
{
\includegraphics[width=0.3\linewidth]{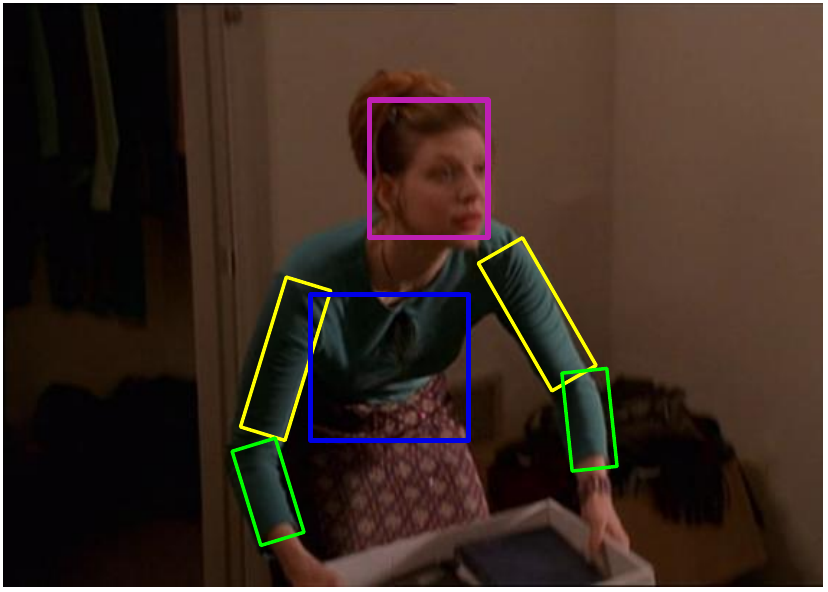}
}
\subfloat[]
{
\includegraphics[width=0.3\linewidth]{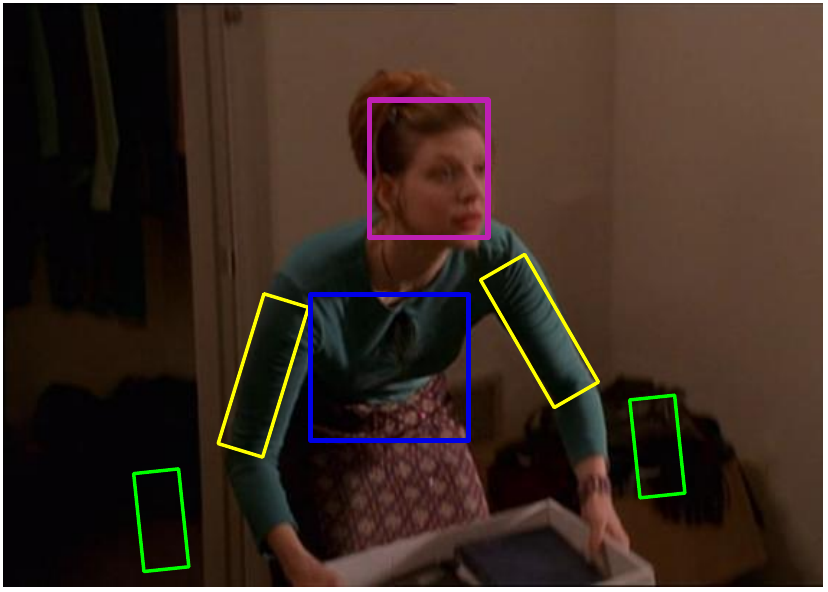}
}
\subfloat[]
{
\includegraphics[width=0.3\linewidth]{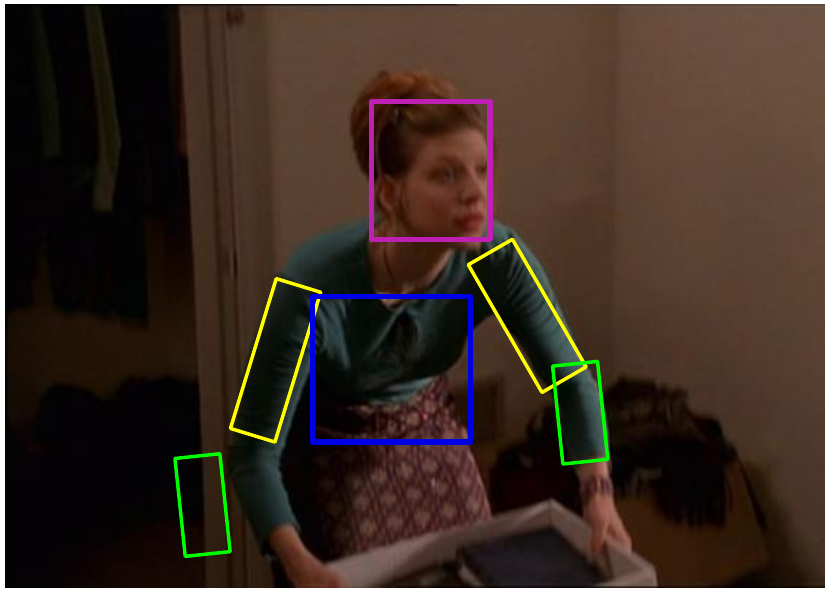}
}
\caption{ \textbf{Examples to demonstrate the benefit of integrating clothing attributes into HPE.}
In the three results of HPE, all human poses in (b) and (c) are correct except lower arms.
we can assume that (c) is incorrect based on the great appearance difference between left and right lower arm,
but there is slight appearance difference in (b).
If we know the clothing attribute type, e.g. the sleeve type or color,
we can remove (b) based on the inconsistent color between the upper and lower arms.
Finally, we get the correct estimation (a).
}
\label{fig:eg}
\end{figure}

To tackle the challenges, it is well-recognized that the contextual information (e.g., clothing attributes) is useful, as illustrated in Figure~\ref{fig:eg}. As a consequence, the so-called \emph{context modeling}, which is to model properly the contextual information possibly existing in images, is widely regarded as a promising direction for HPE. A variety of approaches have been proposed and investigated in the literature over several years, e.g.,~\cite{songchun,ladicky,shen2014unified}. In~\cite{songchun}, it was proposed a model that encourages high contrast between background and foreground. Ladicky et al.~\cite{ladicky} combined together pose estimation and image segmentation, aiming to take the advantages of joint learning. In~\cite{shen2014unified}, a unified structured learning procedure was adopted to predict human pose and garment attribute simultaneously.

While effectual, the existing approaches require to label lots of contextual messages for training, and thus they are time-consuming and impractical. In this paper, we shall introduce a \emph{latent} clothing attribute approach for HPE. Our approach formulates the HPE problem by extending the pictorial structure framework~\cite{ps1,ps2} and, in particular, models the clothing attributes as \emph{latent variables}. Comparing to the previous approaches that rely on label information, our latent approach, in sharp contrast, requires no explicit labels of the clothing attributes and can therefore be executed in an efficient way. We define some clothing attributes and build their connections with human parts (e.g., sleeve with arms). Some domain specific features, including \emph{pose-specific} features and \emph{pose-attribute} features, are designed to describe the connections. We utilize the latent structured support vector machines (LSSVM) for the training procedure, where the attribute values are initialized by a simple K-Means clustering algorithm. Then the model parameters are learnt by employing a relabel strategy, which minimizes the objective function of LSSVM in an ``alternating direction'' manner. More precisely, we perform an iterative scheme to train the model: Given the (latent) clothing attributes, we perform a dynamic programming algorithm to find a suboptimal solution for human pose; Given the human pose, we seek the optimal attribute values by performing a greedy search on the attribute space. We empirically show that our approach can achieve the state-of-the-art performance on two benchmarks.

In summary, the contributions of this paper are three-folds: (1) We establish a latent clothing attribute approach that can implicitly utilize clothing attributes to enhance HPE. (2) We propose some domain specific features to describe the connections between human parts and clothing attributes. (3) We introduce an efficient algorithm to solve the optimization problem which is indeed challenging due to the presence of latent variables.

\section{Related Work}

As aforementioned, HPE is a difficult problem, especially in unconstrained scenes. Some of the researchers studied the problem under the context of 3D scenery~\cite{burenius20133d,ics14cvpr}. In the work of~\cite{burenius20133d}, they extended the popular 2D pictorial structure~\cite{ps1,ps2} to 3D images and employed the new framework to model view point, joint angle, etc. Shotton et al.~\cite{shotton2013real} proposed a real time algorithm for estimating the 3D human pose, striving for making the technique practical in real world applications.

Most studies (including this work) on HPE focus on 2D static images. In the early works, the human part was often modeled by oriented template. Although straightforward, the oriented templates may not properly handle the fore-shortening of the objects~\cite{nips06,cvpr10,Daniel}. In~\cite{deva11}, an advanced representation scheme was proposed to model the oriented human parts. The new model is formulated as a mixture of non-oriented components, each of which is attributed with a ``type''. Interestingly, the new model can approximate the fore-shortening by tuning the adjacent components in a spring structure.

Some work tried to incorporate ``side'' techniques, e.g., image segmentation, to enhance HPE. In~\cite{eccv10}, a variety of image features, e.g., boundary response and region segmentation, were utilized to produce more reliable HPE results. In~\cite{songchun}, the background was modeled as a Gaussian distribution. In~\cite{mixing}, the authors present a two-stage approximate scheme to improve the accuracy of estimating lower arms in videos. The algorithm was imposed to output the candidates with high contrast to the surroundings.

Besides of the shape feature which is very discriminative, the appearance feature (e.g. color, texture) is also important for HPE~\cite{bmvc09}. Generally, the appearance feature is actually a description of the clothing. As illustrated in Figure~\ref{fig:eg}, there is a strong correlation between human pose and clothing attribute. Some previous work such as~\cite{clotheccv,clothliu,poselets,junchi} utilized the result of HPE to predict the clothing attribute or retrieve similar garments. Other methods (e.g.,~\cite{cloth12,shen2014unified}) attempted to refine the clothing parsing by HPE and, in turn, refine HPE by clothing parsing. However, this requires a large annotation for clothing. In our work, it is not required to manually annotate the attributes as we take them as latent variables.

There is some work that has investigated clothing attributes in the tasks other than HPE. In~\cite{clothrec}, Liu et al. aimed to recommend garment for specific scenes. To bridge the gap between the low-level image evidence and the garment recommendation, they integrated an attribute-level representation that propagates semantic messages to the recommendation system. In~\cite{action}, similar attribute techniques as ours were used for action recognition. However, there is a key difference: In~\cite{action}, the attribute is used as a middle level prior and the high level task was facilitated by the knowledge of attribute; In our work, the attribute is modeled in a unified manner with human pose. Our model takes a relabel strategy to alternatively optimize the variables of the attribute and pose.

\begin{figure}[tbp]
\centering
\includegraphics[width=\textwidth]{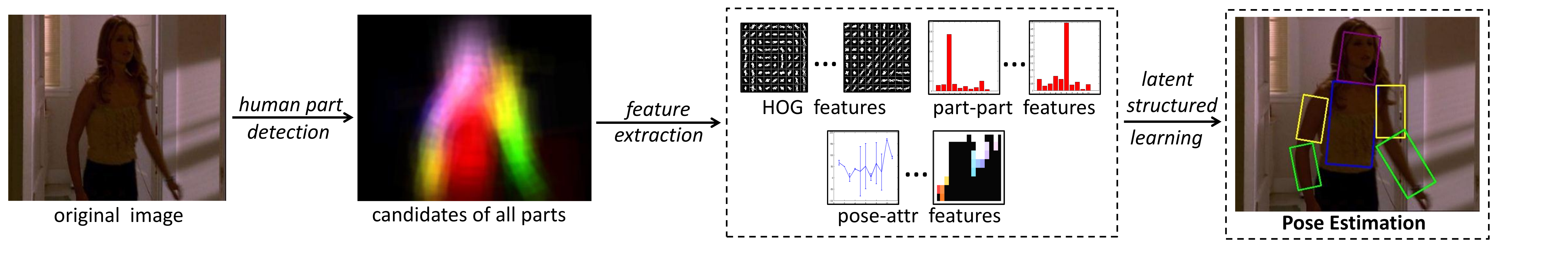}
\caption{ \textbf{Overview of our approach.} }
\label{fig:frame}
\end{figure}

\section{HPE with Latent Clothing Attributes}
We summarize the pipeline of our approach in Figure~\ref{fig:frame}. First, we take a pre-processing step to detect potential human parts in the image. This step allows us to have a search space with manageable size. Then, we extract the domain specific features to characterize the human pose and clothing attributes. Finally, we utilize the LSSVM to actualize our attribute aware human pose model and present an efficient inference algorithm to find an approximate optimal solution to LSSVM. Note that our model can reveal the clothing attributes, and thus humans with similar attribute values will be grouped together (i.e., clustering human by their clothing attributes).

\begin{table}
\centering
\caption{The configuration of clothing attributes}
\begin{tabular}{|c|c|c|c|} \hline
Attribute & Human parts & Features & Number of values \\ \hline
Sleeve &  All arms & Color Histogram & 3 \\ \hline
Neckline & Torso + Head & HOG & 4\\ \hline
 Pattern & Torso & LBP~\cite{lbp} & 5\\ \hline
\end{tabular}
\label{tb:attr}
\end{table}

Before introducing the proposed approach in detail, we would like to introduce some notations. We write $I$ for an image. A human part is represented as a bounding box $(x, y, s, \theta)$, where $(x, y)$ is the coordinate, $s$ is the size and $\theta$ is the rotation. To obtain an input space with manageable size, we use the existing HPE method~\cite{deva11} to produce 40 candidates for each human part. Thus, the input space $\mathcal{X}$ of our approach is defined as:
\begin{equation}
    \mathcal{X} = \left\{ \mathbf{x}|\mathbf{x} = ( \mathbf{b}_1, \mathbf{b}_2, \cdots, \mathbf{b}_m ) \right\},
\end{equation}
where $m$ is the number of human upper-body parts ($m = 6$ in this work), and $\mathbf{b}_i$ denotes the candidate ensemble for the $i$-th human part (there are 40 candidates in each $\mathbf{b}_i$). The output space of human pose is defined as as follows:
\begin{equation}
    \mathcal{P} = \{\mathbf{p}| \mathbf{p}=(p_1,p_2,\cdots,p_m), \forall i, 1\leq p_i \leq 40\},
\end{equation}
where $p_i$ is a positive integer that indicates the index of the estimated candidate.

We aim to integrate clothing attributes into HPE task, striving for capturing the strong correlation between human parts and clothing attributes. We consider three types of attributes in this work, including ``Neckline'', ``Pattern'' and ``Sleeve''. Each attribute has multiple styles, e.g., short sleeve and long sleeve for the ``Sleeve'' attribute. Heuristically, for each $r$-th attribute ($r=1,2,3$), the number of attribute values, $T_r$, are determined as in Table~\ref{tb:attr} (see the last column). Then the output space of the latent clothing attributes is as follows:
\begin{equation}
    \mathcal{A} = \left\{ \mathbf{a}|\mathbf{a} = (a_1, a_2, \cdots, a_n), \forall r, 1 \leq a_r \leq T_r \right\}.
\end{equation}
where $n$ is the number of clothing attributes ($n = 3$ in this work), and $a_r$ is the label for the $r$-th attribute. Note here that it has no specific consideration to choose the value for $a_r$, e.g., $a_1 = 1$ may mean short sleeve or long sleeve. In this work it is an unsupervised clustering procedure that recognizes the clothing attributes.

Finally, the task of jointly estimating clothing attribute and human pose is formulated as follows:
\begin{equation}
    f: \mathcal{X} \rightarrow \mathcal{Y},
    \label{eq:task_func}
\end{equation}
where $\mathcal{Y}$ is the output space given by
\begin{equation}
    \mathcal{Y} = \left\{ \mathbf{y}|\mathbf{y} = (\mathbf{p},\mathbf{a}), \mathbf{p} \in \mathcal{P}, \mathbf{a} \in \mathcal{A} \right\}.
\end{equation}

Regarding the prediction function $f$, we presume that there is a score function $S$ which measures the fitness between any input-output pair $(\mathbf{x}, \mathbf{y})$ such that:
\begin{equation}
\label{eq:S(x,y;beta)}
    S(\mathbf{x}, \mathbf{y};\beta) = \langle \beta, J(\mathbf{x}, \mathbf{y}) \rangle
\end{equation}
where $\langle \cdot \rangle$ denotes the inner product between two vectors, $J(\cdot, \cdot)$ is the feature representation, and $\beta$ is an unknown weight vector. In this way, the mapping function $f$ in Eq.~\ref{eq:task_func} can be written as:
\begin{equation}
    f(\mathbf{x}; \mathbf{\beta}) = \argmax_{\mathbf{y} \in \mathcal{Y} } S(\mathbf{x}, \mathbf{y}; \mathbf{\beta})
    \label{eq:score}
\end{equation}
This is a latent structured learning problem, where the latent variables are clothing attributes. Our learning procedure is motivated by~\cite{dpm}, which employs a relabel strategy to increasingly improve the prediction of latent variables. Yet before proceeding to the training pipeline, we firstly introduce the design of the domain-specific features, as shown in the next section.

\subsection{Feature Representation}
\label{sec:feature}
The joint feature representation is an important component in structured learning~\cite{svm-struct}. We define the joint feature function $J(\mathbf{x}, \mathbf{y})$ by using two types of features, including {\em pose-specific} features denoted by $j_p(\mathbf{x}, \mathbf{p})$,
and {\em pose-attribute} features denoted by $j_{pa}(\mathbf{x}, \mathbf{y});$ that is,
\begin{equation}
    \label{eq:feature}
    \langle \mathbf{\beta}, J(\mathbf{x},\mathbf{y}) \rangle = \langle \mathbf{\beta}_p, j_p(\mathbf{x}, \mathbf{p}) \rangle + \langle \mathbf{\beta}_{pa}, j_{pa}(\mathbf{x},\mathbf{y}) \rangle
\end{equation}
In the following, we present our techniques used to design each type of feature.

\subsubsection{Pose-specific Features}
Given an input sample $\x$, we use the Histogram of Oriented Gradients (HOG)~\cite{hog} to describe the shape of a candidate and consider the deformation constraint between two connected parts:
\begin{equation}
    j_p(\mathbf{x}, \mathbf{p}) = \sum_{i=1}^m hog(\mathbf{x}, p_i) + \sum_{(i, j) \in E_p} d(\mathbf{x}, p_i, p_j),
\end{equation}
where $E_p$ is the set of connected limbs.
The design of the deformation feature $d(\mathbf{x}, p_i, p_j)$ involves some basic geometry constraints between connected parts,
including relative position, rotation and distance of part candidate $p_i$ with respect to $p_j$,
which is computed as $[x_j - x_i, y_j - y_i, (x_j - x_i)^2, (y_j - y_i)^2]$~\cite{deva11}.

\subsubsection{Pose-Attribute Features}
Now we try to integrate the clothing attributes into our model. Notice that an attribute is only associated with some of the human parts). For a given attribute $r$, we denote the human parts associated with it as $r_p$ and the corresponding configuration as $P_r$. The detailed inter-dependency between human parts and clothing attributes is shown in the second column of Table~\ref{tb:attr}. According to the work~\cite{clothliu}, for different attributes, different low-level features should be used to achieve good performance. The specific features used for each clothing attribute can be found in the third column in Table~\ref{tb:attr}.

Formally, the pose-attribute features are defined as:
\begin{equation}
    \label{eq:j_pa}
    j_{pa}(\mathbf{x}, \mathbf{y}) = \sum_{r=1}^n \Psi(\mathbf{x}, P_r, a_r)
\end{equation}
where $\Psi(\mathbf{x}, P_r, a_r)$ denotes the features extracted from the human part $\textbf{x}$, with the configuration $P_r$ and the attribute label $a_r$.

\begin{algorithm}
\caption{Structured Learning with Latent SVM}
\begin{algorithmic}[1]
    \REQUIRE Positive samples, negative samples, initial model $\beta$, number of relabel iteration $t_1$, number of hard negative mining iteration $t_2$.
    \ENSURE Final Model ${\beta}^*$.
    \STATE Initialize the final model: ${\beta}^* = \beta$.
    \STATE Let the negative sample set $F_n = \emptyset$.
    \FOR{ relabel = 1 to $t_1$ }
        \STATE Let the positive sample set $F_p = \emptyset$.
        \STATE Add positive samples to $F_p$.
        \FOR{ iter = 1 to $t_2$ }
            \STATE Add negative samples to $F_n$.
            \STATE ${\beta}^* := \mathrm{Pegasos}({\beta}^*, F_p \bigcup F_n)$.\\
            \STATE Remove easy negative samples: \\
             Remove the samples whose feature vector $v$ satisfying $\langle{\beta}^*, v \rangle < -1$ from $F_n$.
        \ENDFOR
    \ENDFOR
\end{algorithmic}
\label{alg:train}
\end{algorithm}

Similar to~\cite{shen2014unified}, the pose-attribute feature is designed by an outer product of low-level features and an identity vector. We first convert the clothing attribute label $a_r$ to a $T_r$-dimensional vector, denoted as $L(a_r)$, one element of which is assigned with valued ``1'' and all others are set to be ``0''. From Table~\ref{tb:attr}, the low-level feature descriptors of the $r$-th clothing attribute depend on two aspects:
1) the corresponding human parts and 2) the feature type (denoted by $F_r$ and has been specified in Table~\ref{tb:attr}). We use $F_r(P_r)$ to denote features of the $r$-th clothing attribute associated with the part configuration $P_r$. Then our pose-attribute feature $\Psi(\mathbf{x}, P_r, a_r)$ is designed as follows:
\begin{equation}
    \Psi_{pa}(\mathbf{x}, P_r, a_r) = F_r(P_r) \otimes L(a_r)
\end{equation}
where the ``$\otimes$'' operator represents the (vectorized) outer product of two vectors.
\subsection{Structured Learning with Latent SVM}
Now we consider the problem of learning the prediction mapping $f$, given a collection of images labeled with human part locations. This is the type of data available in the all standard benchmark dataset for human pose estimation. Note that clothing attributes have no labels, and we treat them as latent variables.

We describe a framework for initializing the structure of a joint model and learning all parameters. Parameter learning is done by constructing a LSSVM training problem.
We train the LSSVM using the relabel approach (details will be described later) together with the data-mining (hard negative mining), and we use Pegasos~\cite{pegasos} for the online update to solve the problem of huge space for negative samples.

\begin{algorithm}
\caption{Inference for Clothing Attributes}
\begin{algorithmic}[1]
    \REQUIRE A sample $\mathbf{x}$, Model parameter $\beta$ , Human parts label $\mathbf{p}$
    \ENSURE optimal clothing attributes value $\mathbf{a^*}$
    \STATE let $T_r$ is the number of $r$-th clothing attribute type
    \FOR {r:= 1 \textbf{to} 3}
        \STATE select the attribute value which has highest score:\\
            $\mathbf{a}_r = \argmax_{1 \leq r \leq T_r} \langle \beta_{pa}^r, j_{pa}(\mathbf{x}, P_r, a_r) \rangle $
    \ENDFOR
\end{algorithmic}
\label{alg:attr}
\end{algorithm}

\subsubsection{Objective Function}
We aim to learn the fitness function $S(\x, \y; \beta)$ defined in Eq.~\eqref{eq:S(x,y;beta)}, which can later be used for joint estimation (see Eq.~\eqref{eq:score}). Given a positive training sample $(\x, \y)$, we expect $S(\x, \y; \beta) \geq 1$. On the other hand, if a training sample $(\x, \y)$ is negative, the output of the fitness function is required to be less than $-1$. In this way, given a training set $D = \{ (\x_1, \y_1, z_1), \cdots, (\x_q, \y_q, z_q) \}$, where $z_k \in \{1, -1\}$ indicates the $k$-th sample is positive or not, we can optimize the following objective function to solve $\beta$:
\begin{equation}
\begin{split}
\min_{\beta}  \frac{1}{2} \|\beta\|^2 + C \sum_{k=1}^{q}\max(0, 1 - z_k S(\x_k, \y_k; \beta)).
    \label{eq:latent}
\end{split}
\end{equation}

\subsubsection{Initialization}
Since the clothing attributes are latent variables, we can only access the label of human pose. To start up, we take a relabel strategy to update the positive samples (more accurately, the clothing attribute labels) and the weight vector $\beta$ in an alternative manner.

There are many ways to initialize the latent variables. One can randomly assign labels for training samples which may be unstable. In our work, we first use the groundtruth of human pose to extract low-level features (see Table~\ref{tb:attr}) for each attribute. Then we perform a $K$-Means clustering algorithm to obtain the center of each attribute value, where $K$ is exactly the number of attribute values we defined in Table~\ref{tb:attr}. In this way, the initial label for the clothing attribute can be determined by the closest center.

Now all of the labels have been generated, we can solve Problem~\eqref{eq:latent} to obtain the initial weight vector $\beta$ (line 1 in Algorithm~\ref{alg:train}).

\subsubsection{Relabel Strategy}
As the initial clothing attribute labels are not accurate, we employ a relabel strategy to update the attribute labels. That is, given the model parameter $\beta$ and human pose, we predict the clothing attribute by maximizing the fitness function $S(\x, \y; \beta)$, which is shown in Algorithm~\ref{alg:attr}. Note that according to the design of our joint feature $J(\x, \y)$, the pose-specific features are irrelevant for the inference of attributes. From Eq.~\eqref{eq:j_pa}, we know that there is no interaction between different attributes since $j_{pa}$ is summation of $n$ separate attributes associated features. Therefore, we can perform an efficient greedy search for each attribute to obtain a local optima (line 2--4 in Algorithm~\ref{alg:attr}).

\begin{algorithm}
\caption{Approximate Inference for Clothing Attribute Aware HPE Task}
\begin{algorithmic}[1]
    \REQUIRE A sample $\mathbf{x}$, Model parameter $\beta$.
    \ENSURE Optimal estimation $\mathbf{y}^*$ and score $S*$.
    \STATE Set $\mathbf{y}^* = \emptyset$.
    \STATE Set the optimal score $S^* = -\infty$.
    \STATE Initialize the parts estimation $\mathbf{p}_0$.
    \REPEAT
        \STATE Compute the local optimal clothing attributes $\mathbf{a}_t$.
        \STATE Compute the local optimal human pose $\mathbf{p}_t$.

        \STATE Compute the local score: $S = S(\mathbf{x}, \mathbf{y}_t; \beta)$.

        \IF{$S > S^*$}
            \STATE $S^* = S$, $\mathbf{y}^* = \mathbf{y}_t $
        \ENDIF
    \UNTIL{$S^*$ not change}
\end{algorithmic}
\label{alg:inference}
\end{algorithm}

\subsubsection{Hard Negative Mining}
For a recognition or detection task, one can obtain a positive sample set with manageable size. However, there is a huge space for the negative samples. Actually, it is not possible for enumerate \emph{all} negative samples. Thus, it is important to feed an algorithm with ``hard'' negative samples for efficiency and memory cost. In line 6--10 of Algorithm~\ref{alg:train}, we perform hard negative mining~\cite{dpm} to obtain valuable negative samples. This schema will call the inference algorithm~\ref{alg:inference} (see Section~\ref{subsec:inference}). More concretely, given an input sample $\x$ and weight vector $\beta$, we launch Algorithm~\ref{alg:inference} to find the optimal estimation $\y^*$. If $z \cdot S^*$ is less than $-1$ (a threshold we set), $\x$ is considered hard. The searching procedure on $\x$ will be stopped only when the $S^*$ is greater than $-1$ (the $\y^*$ produced by the previous step is removed from the search space).

After collecting all the hard negative samples, we update $\beta$ with Pegasos solver~\cite{pegasos} (line 8 in Algorithm~\ref{alg:train}). Then we use the updated $\beta$ to perform a shrinkage step to remove the easy negatives from the hard negative set $F_n$.

\begin{figure}[tbp]
\centering
\includegraphics[width=0.9\textwidth]{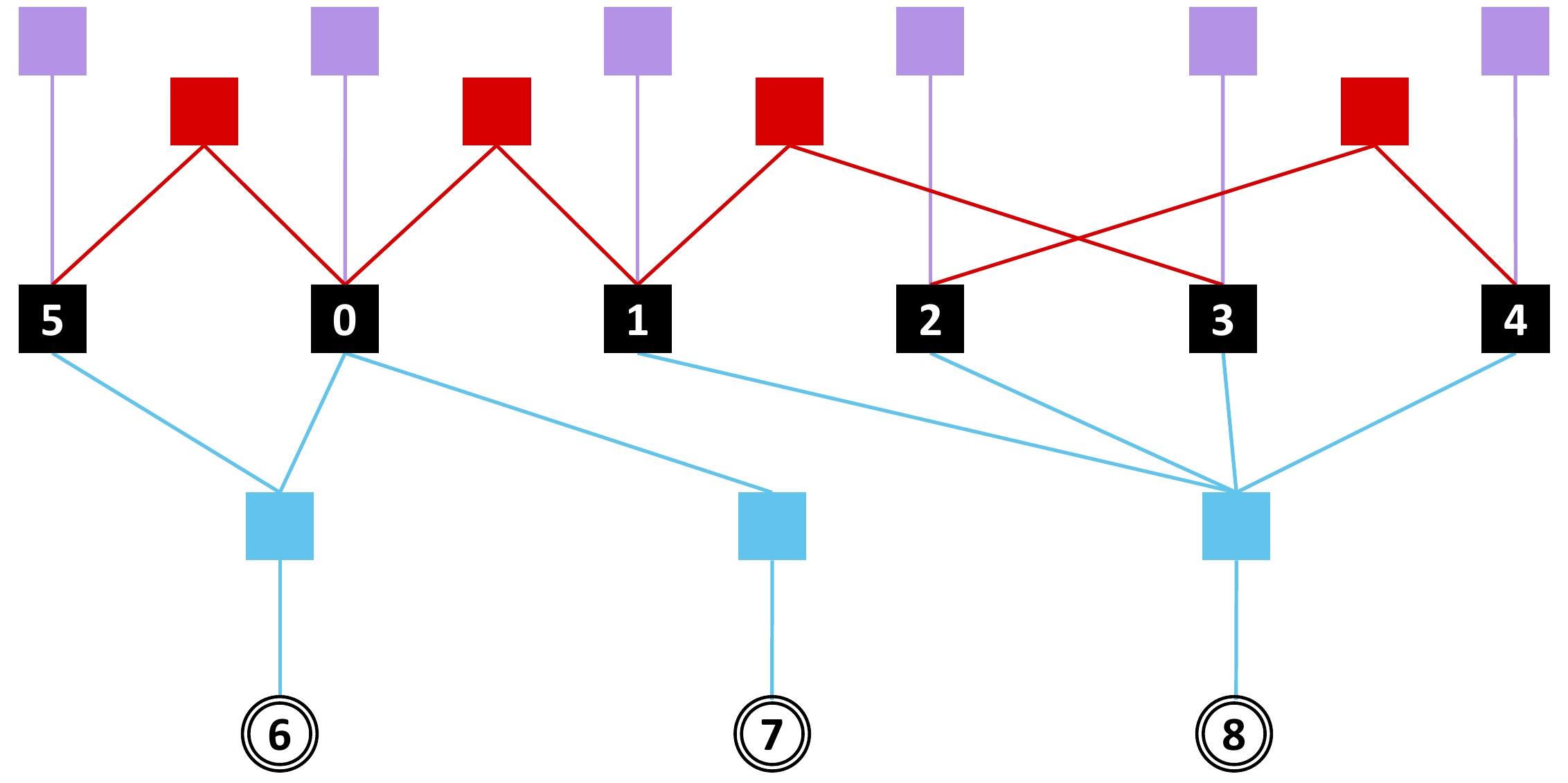}
\caption{ \textbf{Nodes with numbers from 0 to 5 are the human part variable and those 6 to 8 are clothing attributes. Colored nodes are the potentials.} }
\label{fig:graph}
\end{figure}

\subsection{Inference}
\label{subsec:inference}
In Figure~\ref{fig:graph}, we represent our problem as a factor graph $\mathcal{G}$, where the rectangle node denotes a human part,
the circle node with double boundaries denotes a clothing attribute.
As our original problem is a cyclic graph, it cannot be optimized exactly and efficiently.
Therefore, in Algorithm~\ref{alg:inference}, we propose an iterative algorithm to search for an approximate solution.
Our algorithm receives a sample x, the model parameter $\beta$ as inputs and outputs a local optima for human parts and clothing attribute.
In each iteration, by fixing the attributes, the inference can be performed on a tree structure, which can be optimized with a dynamic programming~\cite{ps2}. When the human parts are fixed, an efficient greedy search schema for clothing attribute is employed (see Algorithm~\ref{alg:attr}).

\begin{algorithm}
\caption{Inference for Human Pose}
\begin{algorithmic}[1]
    \REQUIRE A sample $\mathbf{x}$, Model parameter $\beta$ ,  Clothing attributes value $\mathbf{a}$
    \ENSURE optimal human parts estimation $\mathbf{p^*}$
    \STATE set the optimal human parts estimation $\mathbf{p^*} = \emptyset$
    \STATE set the node 0 as the root node
    \FOR{ each candidate $\mathbf{p}_i$ of node $i$ }
        \STATE set $m(\mathbf{p}_i) = \langle \beta_p^i, \phi_p(\mathbf{x}, p_i) \rangle + \langle \beta_{pa}^r, \Psi_{pa}(\mathbf{x}, P_r, a_r) \rangle$
    \ENDFOR
    \FOR{ each candidate $\mathbf{p}_j$ of parent node $j$ and $\mathbf{p}_i$ of child node $i$ }
        \STATE set $l(\mathbf{p}_i, \mathbf{p}_j) = \langle \beta_p^{ij}, \psi_p(\mathbf{x}, p_i, p_j) \rangle$
        \IF{ $i$ is a leaf node }
            \STATE $B_i(\mathbf{p}_j) = \max_{\mathbf{p}_i} (m(\mathbf{p}_i) + l(\mathbf{p}_i, \mathbf{p}_j) )$
        \ELSE
            \STATE $B_i(\mathbf{p}_j) = \max_{\mathbf{p}_i} (m(\mathbf{p}_i) + l(\mathbf{p}_i, \mathbf{p}_j) + \sum_{v \in C_i} B_v(\mathbf{p}_i) )$
        \ENDIF
    \ENDFOR
    \STATE select the best candidate for the root node: \\
        $\mathbf{p}_0^* = \arg \max_{\mathbf{p}_0} ( m(\mathbf{p}_0) + \sum_{v \in C_0} B_v(\mathbf{p}_0) )$
    \FOR{ each parent-child pair ($\mathbf{p}_j^*, \mathbf{p}_i$) }
        \STATE $\mathbf{p}_i^* = \arg \max_{\mathbf{p}_i} B_i(\mathbf{p}_j^*)$
    \ENDFOR
\end{algorithmic}
\label{alg:ps}
\end{algorithm}

\subsubsection{Inference for Human Pose}
We elaborate the inference procedure of human pose by extending the pictorial structure framework.
In Figure~\ref{fig:graph}, we denote our score with colored nodes,
with purple and red ones denoting the appearance and deformation scores.
The main extension for the traditional PS model is the cyan nodes,
which denoting the score to measure the fitness of human pose and clothing attribute (called pose-attribute score).
Therefore, we propose the human pose inference procedure in Algorithm~\ref{alg:ps}.
We denote the children nodes as $C_i$ for a node $i$.
We compute the appearance and pose-attribute scores in line 3--5.
In line 7, we compute the deformation score for each parent-child pair node $i$ and $j$.
In the line 8--12, we compute conventional message passing procedure by dynamic programming~\cite{ps1}.
Then we perform a top-down process to find the best candidate for each human part in line 14--17.

\section{Experiments}
\label{sec:exp}

\subsection{Datasets}
We evaluate our approach using the Buffy dataset~\cite{fer08} and the DL (daily life) dataset.
The Buffy Dataset contains 748 pose-annotated video frames from Buffy TV show.
This dataset is presented as a benchmark for HPE task.
The DL dataset contains 997 daily life photos collected from the Flickr website.
We annotate the human pose for this dataset.
Compared with Buffy, the DL dataset has more various clothing attribute values.
In order to obtain quantitative evaluation results for attributes, we manually annotate the clothing attributes for Buffy and DL.
There is a standard partition of Buffy for training and testing, where the training set consists of 472 images and the remaining are used for testing.
For the DL dataset, we select randomly 297 images for training and use the remaining 700 images for testing.

\begin{table}
\centering
\caption{Comparison with State-of-the-art Algorithms on the Buffy Dataset}
\begin{tabular}{|c|c|c|c|c|c|} \hline
    Method & Torso & Upper arms & Lower arms & Head & Total \\ \hline
Andriluka et al.~\cite{cvpr09} &  90.7 & 79.3 & 41.2 & 95.5 & 73.5 \\ \hline
Sapp et al.~\cite{eccv10} & \textbf{100} & 95.3 & 63.0 & 96.2 & 85.5 \\ \hline
Yang and Ramanan~\cite{deva11} & \textbf{100} & 96.6 & 70.9 & \textbf{99.6} & 89.1 \\ \hline
Our Approach & \textbf{100} & \textbf{97.1} & \textbf{78.4} & 99.1 & \textbf{91.6} \\ \hline
\end{tabular}
\label{tb:buffy}
\end{table}

\begin{table}
\centering
\caption{Comparison with State-of-the-art Algorithms on the DL Dataset}
\begin{tabular}{|c|c|c|c|c|c|} \hline
    Method & Torso & Upper arms & Lower arms & Head & Total \\ \hline
Andriluka et al.~\cite{cvpr09} &  97.0 & 91.7 & 84.5 & 94.0 & 90.6 \\ \hline
Sapp et al.~\cite{eccv10} & \textbf{100} & 88.5 & 78.0 & 87.6 & 86.8 \\ \hline
Yang and Ramanan~\cite{deva11} & 99.8 & 95.7 & 87.5 & 95.6 & 93.6 \\ \hline
Our Approach & \textbf{100} & \textbf{97.2} & \textbf{91.3} & \textbf{99.1} & \textbf{95.7} \\ \hline
\end{tabular}
\label{tb:dl}
\end{table}

\subsection{Baselines and Metric}
We compare our approach with three state-of-the-art algorithms:
Andriluka et al.~\cite{cvpr09}, Sapp et al.~\cite{eccv10}, Yang and Ramanan~\cite{deva11}.
For the HPE results, we evaluate them with a standardized evaluation protocol based on the probability of correct pose (PCP)~\cite{fer09},
which measures the percentage of correctly localized human parts.
For the clothing attributes results, we evaluate them with a standardized metric (F1 score) of clustering task.
We use the $K$-Means clustering results as our  baseline for clothing attributes.
First we use the groundtruth of human pose to obtain the clustering center for each attribute value.
Then we perform $K$-Means clustering under a given pose, which is produced by either the state-of-the-art HPE algorithms or the groundtruth.

\begin{figure}[tbp]
\centering
\includegraphics[width=0.9\textwidth]{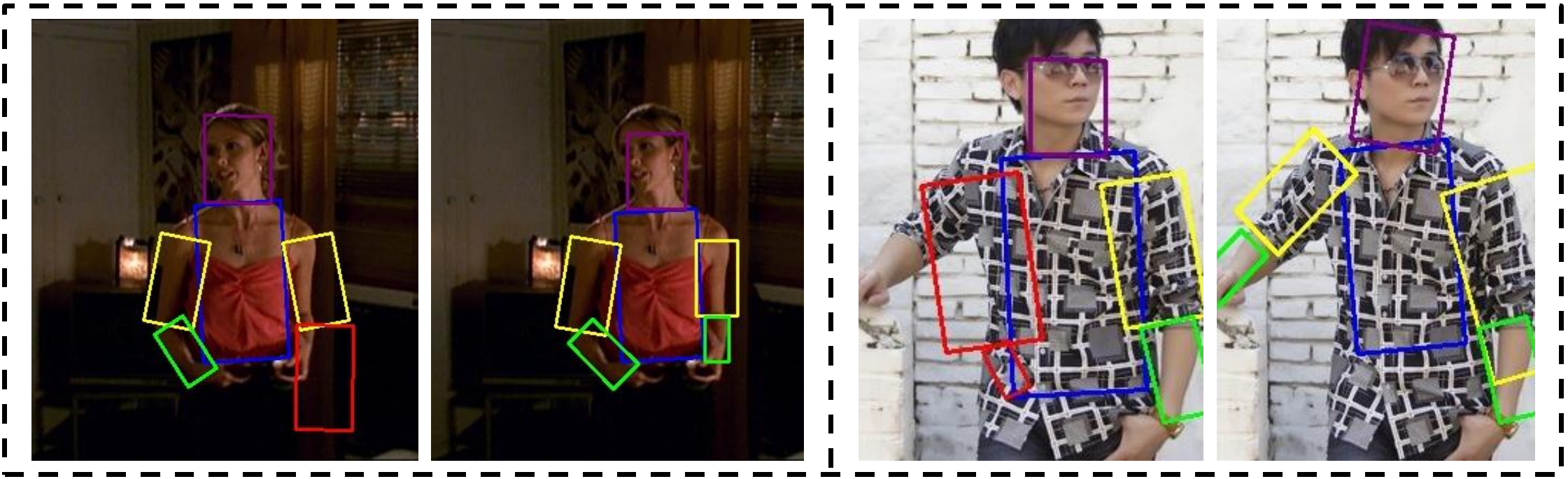}
\caption{ \textbf{Comparison of our approach with Yang and Ramanan~\cite{deva11} }
Yang and Ramanan~\cite{deva11} produces incorrect estimation (the 1st and 3rd) for upper and lower arms,
while our latent clothing attribute approach produces correct.
}
\label{fig:compare}
\end{figure}

\subsection{Results}
\begin{figure}
\centering
\includegraphics[width=\textwidth]{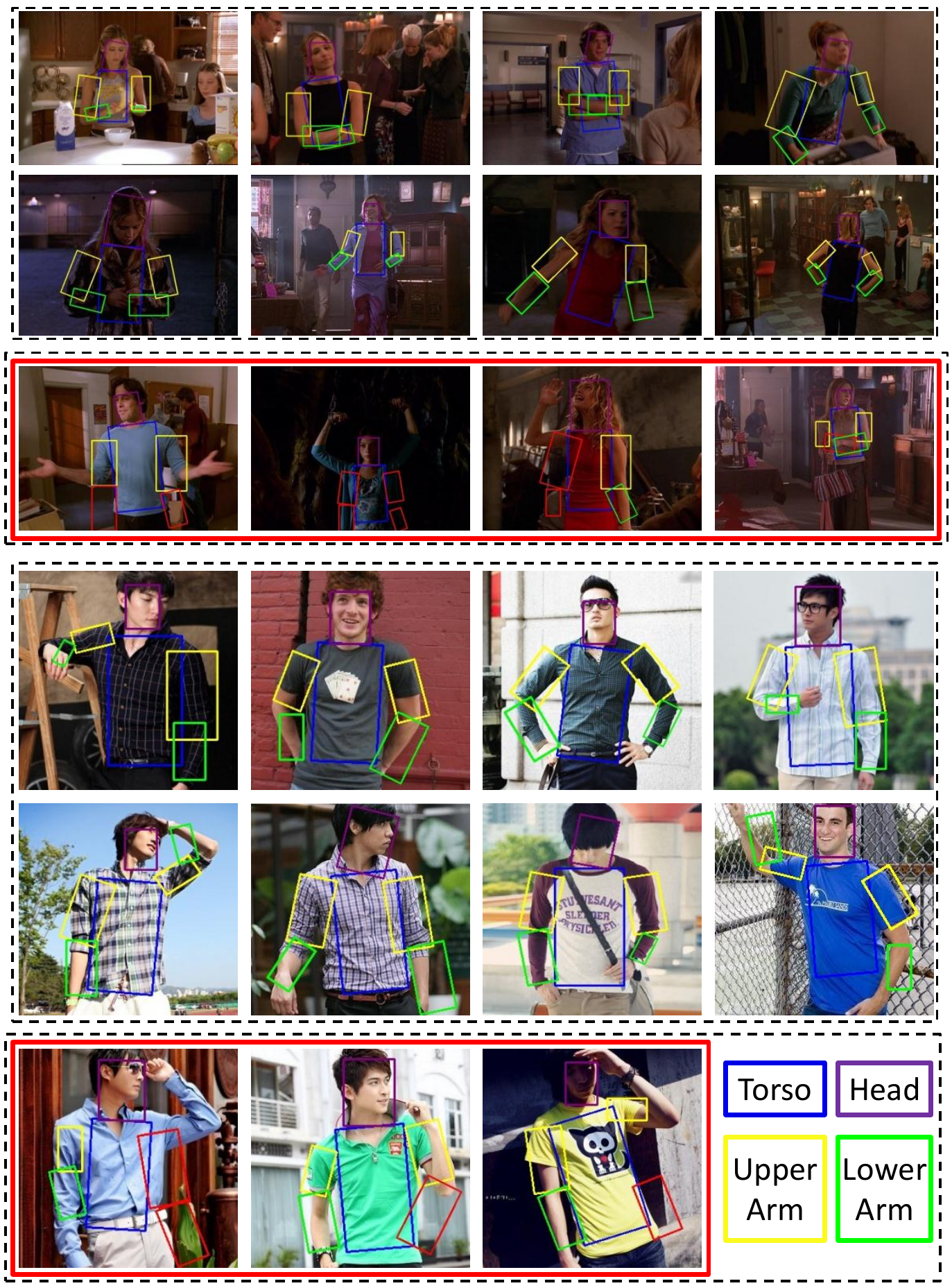}
\caption{ \textbf{Visualization of pose results produced by our algorithm on the Buffy and DL datasets.}
The top two panels are from Buffy and the others are from DL. We use the oriented bounding box to denote the pose estimation.
The first panel of each dataset are correct results, while the second panel are incorrect results.
The bounding box with red color denote the incorrect estimation.}
\label{fig:result}
\end{figure}

Figure~\ref{fig:result} shows some exemplar HPE results produced by our approach.
We provide the PCP evaluation results on Buffy and DL in Table~\ref{tb:buffy} and Table~\ref{tb:dl} respectively.
For the Buffy dataset,
Table~\ref{tb:buffy} shows that our approach consistently outperforms Yang and Ramanan~\cite{deva11} which is a recently established algorithm.
It is expected that the most difficult parts to estimate are the lower arms.
Surprisingly, the improvement on the lower arms of our approach achieves 7.5 percent higher than Yang and Ramanan, possibly because of the integration of the sleeve attribute.
For the DL dataset, our algorithm consistently outperforms all the competing baselines since the photos in DL are collected from daily life and have richer clothing attributes than Buffy.

\begin{table}
\centering
\caption{F1 scores for clothing attributes results on Buffy}
\begin{tabular}{|c|c|c|c|c|} \hline
    HPE & Sleeve & Neckline & Pattern & Total \\ \hline
Andriluka et al.~\cite{cvpr09} + $K$-Means & 24.1 & 26.6 & 34.2 & 28.3  \\ \hline
Sapp et al.~\cite{eccv10} + $K$-Means & 22.9 & 27.9 & 40.5 & 30.4 \\ \hline
Yang and Ramanan~\cite{deva11} + $K$-Means & 38.3 & 25.7 & 22.6 & 28.9\\ \hline
Groundtruth + $K$-Means & 34.7 & 36.1 & 39.5 & 36.8\\ \hline
Our Approach & \textbf{55.6} & \textbf{68.8} & \textbf{80.8} & \textbf{68.4}  \\ \hline
\end{tabular}
\label{tb:f1_buffy}
\end{table}

\begin{table}
\centering
\caption{F1 scores for clothing attributes results on DL}
\begin{tabular}{|c|c|c|c|c|} \hline
    HPE & Sleeve & Neckline & Pattern & Total \\ \hline
Andriluka et al.~\cite{cvpr09} + $K$-Means & 27.5 & 31.7 & 27.6 & 28.9  \\ \hline
Sapp et al.~\cite{eccv10} + $K$-Means & 34.9 & 30.5 & 23.8 & 29.7 \\ \hline
Yang and Ramanan~\cite{deva11} + $K$-Means & 43.2 & 28.6 & 35.8 & 35.9 \\ \hline
Groundtruth  + $K$-Means & 31 & 29.8 & 26.1 & 28.9 \\ \hline
Our Approach & \textbf{57.2} & \textbf{60.3} & \textbf{74.7} & \textbf{64.1}  \\ \hline
\end{tabular}
\label{tb:f1_dl}
\end{table}

\begin{figure*}[tbp]
\centering
\includegraphics[width=\textwidth]{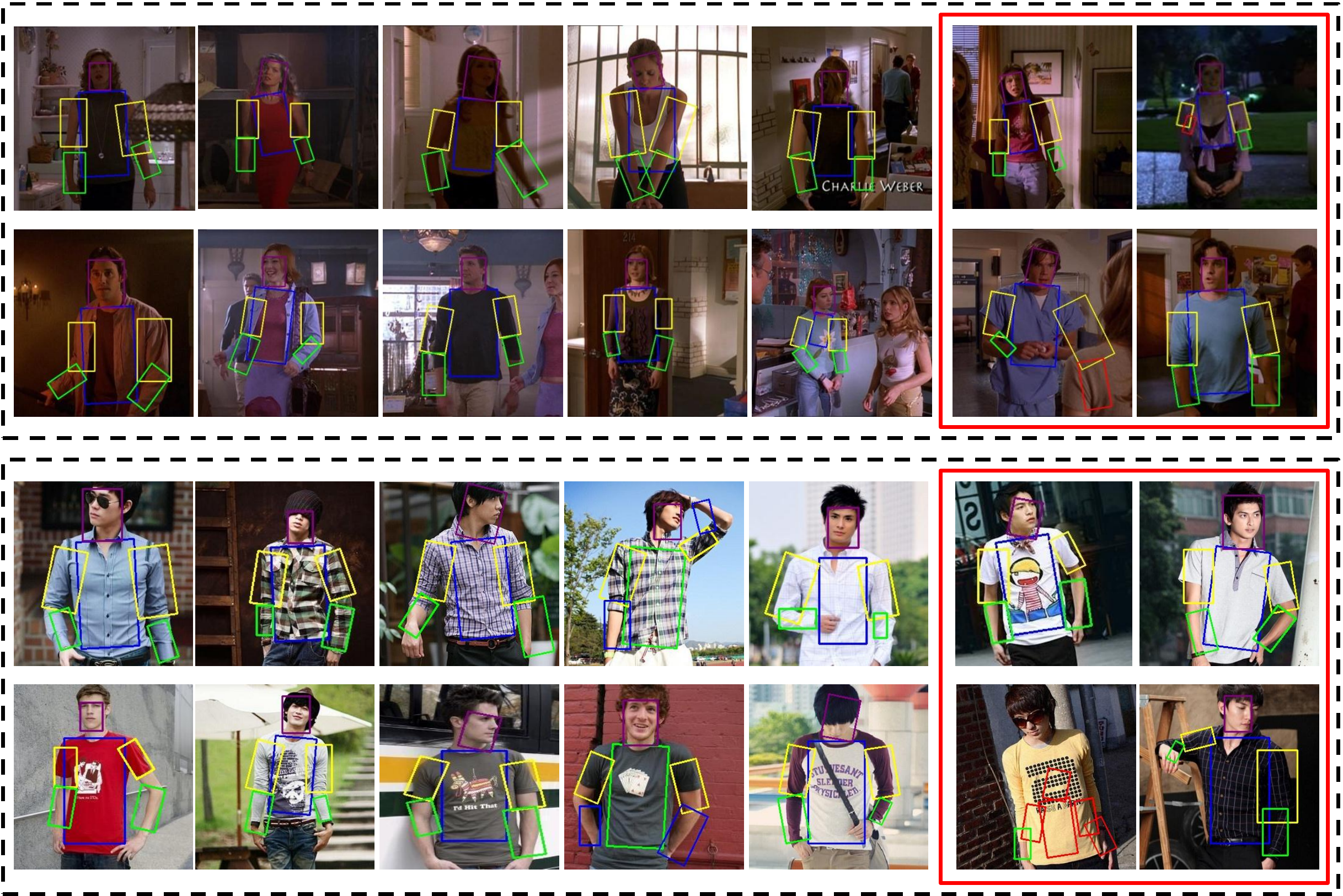}
\caption{ \textbf{Examples grouped on sleeve from Buffy and neckline from DL.}
The first row of the top panel (sleeve) shows the sleeveless type, the second is long type,
while the first row of the bottom panel (neckline) shows the pointed type, the second is round type. The right two columns are the incorrect results.}
\label{fig:sleeve}
\end{figure*}

As we also aim to reveal the clothing attribute, we show some results in Figure~\ref{fig:sleeve} for Buffy and DL, where we arrange the images with same attribute value into one group (i.e. clustering humans by their clothing attributes). In the top pane of Figure~\ref{fig:sleeve}, we group humans by the sleeve attribute. The performance under the F1 score is demonstrated in Table~\ref{tb:f1_buffy} and ~\ref{tb:f1_dl}. Surprisingly, our approach enjoys a significant improvement on both datasets, mainly because of the relabel strategy and the iterative update role for our model parameter. Note that the result of ``$K$-means + Groundtruth'' provides the initial labels for the clothing attributes. In this way, we examine the effectiveness of our relabel strategy.

\section{Conclusion}
\label{sec:conclude}
Inspired by the strong correlation between human pose and clothing attributes, we propose a latent clothing attribute approach for HPE, incorporating the clothing attributes into the traditional HPE model as latent variables.
Compared with previous work~\cite{shen2014unified}, our formulation is more suitable for practical applications as we do not need to annotate the clothing attributes. We utilize the LSSVM to learn all the parameters by employing a relabel strategy. To start up, we take a simple $K$-Means step to initialize the latent variables and then update the model and the clothing attributes in an alternative manner. Finally, we propose an approximate inference schema to iteratively find an increasingly better solution. The experimental results justify the effectiveness of our relabel strategy and show the state-of-the-art performance for HPE.

\bibliographystyle{splncs}
\bibliography{pe_gac}


\end{document}